\documentclass{article}
\usepackage{iclr2027_conference,times}
\usepackage{amsmath,amssymb,amsthm}
\usepackage{booktabs}
\usepackage{graphicx}
\usepackage{multirow}
\usepackage{xcolor}
\usepackage{url}
\usepackage{algorithm}
\usepackage{algorithmic}
\usepackage{pgfplots}
\usepackage{xfp}
\usepackage{siunitx}
\usepackage{hyperref}  % load after algorithm, otherwise table/figure anchors are duplicated
\hypersetup{colorlinks=true,linkcolor=blue!50!black,citecolor=blue!50!black,urlcolor=blue!50!black}
\csname endofdump\endcsname
\usepackage{amsmath,amsfonts,bm}

\def\eqref#1{equation~\ref{#1}}
\def\1{\bm{1}}

\DeclareMathAlphabet{\mathsfit}{\encodingdefault}{\sfdefault}{m}{sl}
\SetMathAlphabet{\mathsfit}{bold}{\encodingdefault}{\sfdefault}{bx}{n}

\pgfplotsset{compat=1.18}
\definecolor{mResolve}{HTML}{2A78D6}
\definecolor{mVerify}{HTML}{EB6834}
\definecolor{mOwnProg}{HTML}{1BAF7A}
\definecolor{mHintProg}{HTML}{EDA100}
\definecolor{mTir}{HTML}{E87BA4}
\definecolor{mOtherProg}{HTML}{008300}
\definecolor{mOtherPool}{HTML}{4A3AA7}
\definecolor{mXeight}{HTML}{E34948}
\newcommand{\pidis}{\pi_{\mathrm{dis}}}
\newcommand{\rhodis}{\rho_{\mathrm{dis}}}

\title{When Does Overriding the Majority Vote Help?}

\author{
Terry Ma$^{1}$ \quad Aidan Zhang$^{1}$ \\[0.5em]
$^{1}$Carnegie Mellon University
}
\iclrfinalcopy

\begin{document}
\maketitle
\lhead{}

%%%%%%%%%%%%%%%%%%%%%%%%%%%%%%%%%%%%%%%%%%%%%%%%%%%%%%%%%%%%%%%%%%%%%%%%%%%%%%%

\begin{abstract}
When should additional reasoning or verification replace a majority-vote answer?
We call such a replacement an override.
We examine two obstacles: the availability of correct alternatives and the reliability of methods that produce or select them.
We compare additional reasoning, executed programs, and cross-model reasoning on 150 competition-mathematics questions with three open reasoning models.
Starting from a majority vote over 16 responses, we measure how often overrides occur and how many errors they correct or introduce across agreement thresholds.
When the vote is incorrect, the correct answer is often absent from the original responses.
Additional same-model reasoning rarely overrides, while executed programs correct more errors than they introduce at every evaluated threshold.
Cross-model overrides improve some weaker baselines but do not establish an advantage over the more accurate model alone, even with only eight standalone responses.
Stricter agreement reduces how often overrides occur but does not consistently increase the fraction that correct errors.
At the strictest thresholds, overrides occur on few distinct questions, limiting conclusions about their reliability.
\end{abstract}

\section{Introduction}

Self-consistency samples many solutions to a problem and returns the most frequent answer~\citep{wang2022selfconsistency}.
The most frequent answer can still be incorrect.
Additional reasoning or verification can correct some of these errors, but it can also replace a correct answer with an incorrect one.
We study when to trust such methods over the majority vote.

Overall accuracy measures the net benefit of these methods, but does not distinguish the errors they correct from those they introduce.
We call a change from the majority-vote answer to a different answer an \emph{override}.
An override is a \emph{rescue} if it replaces an incorrect answer with a correct one, and is a \emph{regression} if it replaces a correct answer with an incorrect one.
Overrides that replace one incorrect answer with another do not affect accuracy.
We report these outcomes separately~\citep{huang2024cannot,mozannar2020consistent}.
Counting rescues and regressions separately shows whether overrides improve accuracy and how often those improvements come at the expense of previously correct answers.

We investigate two obstacles to improving majority voting.
First, is a correct alternative available among the sampled answers?
Second, can an override method identify or produce a correct alternative when it disagrees with the vote?
Our main comparison evaluates additional same-model reasoning, executed programs, and cross-model reasoning for three open reasoning models on the same competition-mathematics questions.
Additional comparisons involving verification, stronger judges, and other methods are reported in the appendix.

Our experiments show that candidate selection is limited by missing correct answers, while methods that produce new answers differ in the rescues and regressions they introduce.
For executed programs, similar accuracy gains at different agreement thresholds conceal differences in the number of regressions.
Cross-model overrides can improve a weaker base, but our comparisons do not establish an advantage over using the more accurate model alone.
We report these outcomes across agreement thresholds rather than selecting a preferred override rule from the observed results.

\section{Evaluating Overrides}
\label{sec:setting}

\subsection{Overview}

\paragraph{Majority vote and candidates.}
We group sampled solutions by their final answers and take the most frequent answer as the majority vote.
We consider two candidate sets: the five most frequent answers and all answers present in the sampled responses.
Candidate availability measures whether the correct answer is in this set: a method restricted to selecting a candidate cannot rescue a wrong vote when the correct answer is absent.

\paragraph{Overrides.}
An override replaces the majority-vote answer with a different answer.
An override method supplies the proposed replacement, using additional reasoning, executed programs, or cross-model reasoning.
The override rule determines when to accept that replacement and when to retain the vote.
For methods that produce several answers, we vary the agreement required before accepting an alternative.

\paragraph{Rescues and regressions.}
An override rescues a case when it replaces an incorrect answer with a correct one, and causes a regression when it replaces a correct answer with an incorrect one.
Replacing one incorrect answer with another leaves accuracy unchanged.
Counting these outcomes separately shows how many errors a method corrects and introduces.

\paragraph{Rescue rate and regression rate.}
The rescue rate is the fraction of overrides that change an incorrect answer to a correct one.
The regression rate is the fraction that change a correct answer to an incorrect one.
These rates need not sum to one: an override can replace one incorrect answer with another.
Comparing these rates tells us whether the overrides improve accuracy.

\paragraph{Override rate.}
The override rate is the fraction of evaluated cases in which the method overrides the vote.
Together with rescue rate and regression rate, it determines the change in overall accuracy.

\subsection{Mathematical definitions}
\label{sec:measure}

Consider $N$ evaluated cases.
Let $O$ be the number of overrides, $R$ the number of rescues, and $D$ the number of regressions.
The remaining $O-R-D$ overrides replace one incorrect answer with another.
The change in accuracy is simply
\begin{equation}
\Delta A = \frac{R-D}{N}.
\label{eq:net}
\end{equation}

Among overrides, the replacement answer is correct in a fraction $R/O$ of cases, while the original vote was correct in a fraction $D/O$.
We call these the \emph{rescue rate} ($\pidis$) and \emph{regression rate} ($\rhodis$), respectively.
The override rate is $O/N$.
For $O>0$, the same accounting can therefore be written as
\begin{equation}
\Delta A =
\underbrace{\frac{O}{N}}_{\text{override rate}}
\left(
\underbrace{\frac{R}{O}}_{\text{rescue rate}}
-
\underbrace{\frac{D}{O}}_{\text{regression rate}}
\right).
\label{eq:intervention}
\end{equation}

Overrides improve accuracy when rescues outnumber regressions.
The size of the improvement also depends on how often the method overrides the vote.
If there are no overrides, the accuracy change is zero and the two rates conditional on overriding are undefined.

These expressions describe equally weighted cases.
Section~\ref{sec:setup} specifies how we aggregate repeated runs and observations from different models, and how we account for their dependence when estimating uncertainty.

\section{Experimental Setup}
\label{sec:setup}

\paragraph{Models and questions.}
We compare three open reasoning models on the same 150 competition-mathematics questions.
The models are DeepSeek-R1-0528-Qwen3-8B (``R1-8B''), gpt-oss-20b, and Qwen3-30B-A3B-Thinking-2507 (``Qwen3-30B'').
The questions come from SMT 2025, CMIMC 2025, HMMT November 2025 and February 2026, and AIME 2026, accessed through MathArena~\citep{balunovic2025matharena}.
Appendix~\ref{app:fixed16} gives the question-selection details and broader evaluations.

\paragraph{Majority-vote baseline.}
We take the majority vote over the final answers extracted from 16 sampled responses.
Responses without a usable final answer do not contribute a vote.

\paragraph{Override rule.}
We override the majority vote only when at least $k$ of eight additional responses agree on a different answer.
Otherwise, we retain the vote.
We evaluate $k\in\{5,6,7,8\}$ for every method; responses without a usable answer provide no support.
This agreement-based rule is similar to the acceptance rule used in LLM cascades~\citep{yue2024large}.

\begin{figure}[htbp]
\centering
\includegraphics[width=\linewidth]{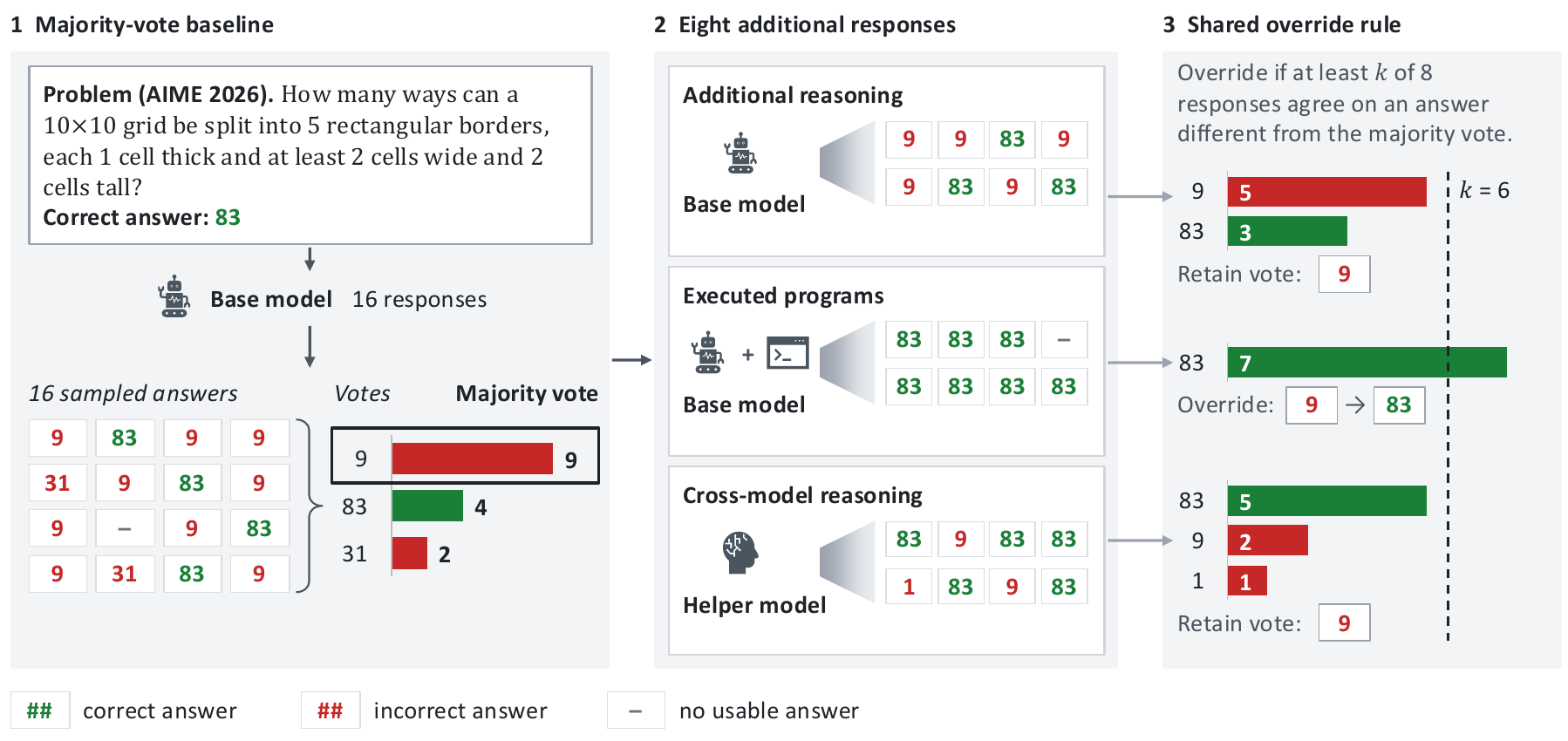}
\caption{Three methods for overriding a majority vote, shown with illustrative responses to an AIME 2026 question.
At $k=6$, seven programs return 83 and override the vote of 9.
Additional reasoning and cross-model reasoning retain the vote.}
\label{fig:override-methods}
\end{figure}

\paragraph{Override methods.}
\label{sec:sources}
We compare additional same-model reasoning, executed programs, and reasoning from a helper model (Figure~\ref{fig:override-methods}).
The base model supplies the original majority vote and generates the additional reasoning responses or programs.
We execute the generated programs to obtain their answers.
In cross-model correction, a helper model supplies the additional reasoning responses.
Appendix~\ref{app:fixed16} documents the criteria for usable responses and program outputs.

\paragraph{Cross-model comparisons.}
We evaluate all six base-helper model pairs, comparing each combined method with both models used alone.
Each standalone model uses the same 16-response majority-vote baseline.
We also compare combined accuracy with majority voting over the exact eight helper responses supplied to the override method.

\paragraph{Statistical analysis.}
We estimate 95\% bootstrap confidence intervals by drawing 30 questions with replacement from each contest in every replicate.
All outcomes for a question are kept together during resampling, including those from different models and repeated runs.
Intervals are calculated separately for each agreement threshold.
Appendix~\ref{app:fixed16} gives the averaging and resampling details.

\section{Results}
\label{sec:results}

\subsection{Candidate availability limits selection}
\begin{figure}[t]
\centering
\includegraphics[width=\linewidth]{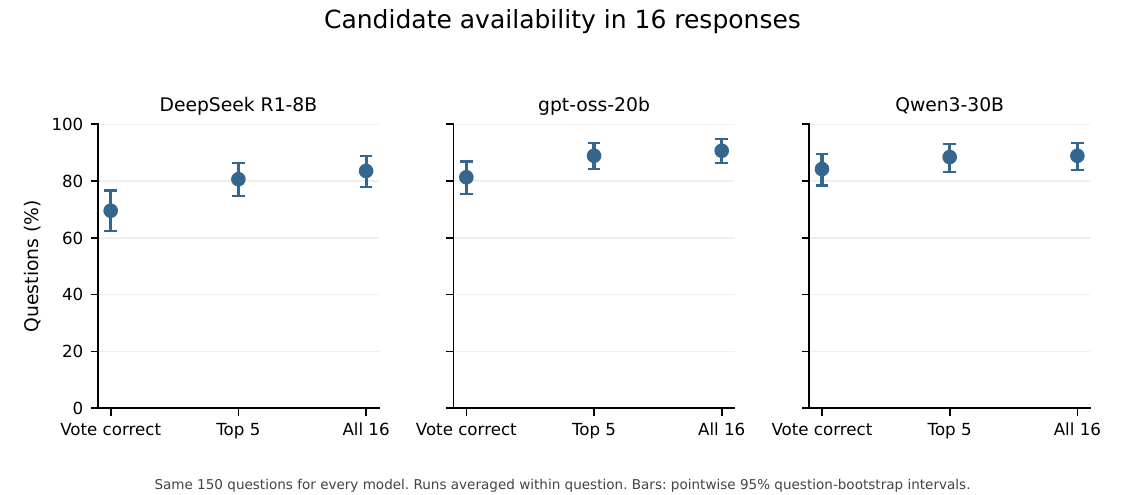}
\caption{\textbf{Majority-vote accuracy and correct-answer availability in 16 responses.}
Each panel compares majority-vote accuracy with the fraction of questions whose correct answer appears among the five most frequent answers or anywhere in the 16 responses.
All models use the same 150 questions, with repeated runs averaged within each question.
Bars show pointwise 95\% question-bootstrap confidence intervals.}
\label{fig:fixed16-availability}
\end{figure}

A method that selects among sampled answers can correct a wrong majority vote only if the correct answer appears in the sample.
Across the three models, majority-vote accuracy is 78.4\%, whereas a correct answer appears among the 16 responses in 87.7\% of cases (Figure~\ref{fig:fixed16-availability}).
Among incorrect votes, the correct answer is present in 43\% of cases (95\% CI [32\%, 56\%]).
Selection from the original responses therefore leaves rescue out of reach for many incorrect votes.

Restricting selection to the five most frequent answers reduces availability slightly in this evaluation.
Methods that generate additional answers can recover solutions absent from the original sample, but those alternatives can also replace correct votes with incorrect answers.
We therefore examine both rescues and regressions.

\subsection{Rescues and regressions differ across methods}
\begin{figure}[t]
\centering
\includegraphics[width=\linewidth]{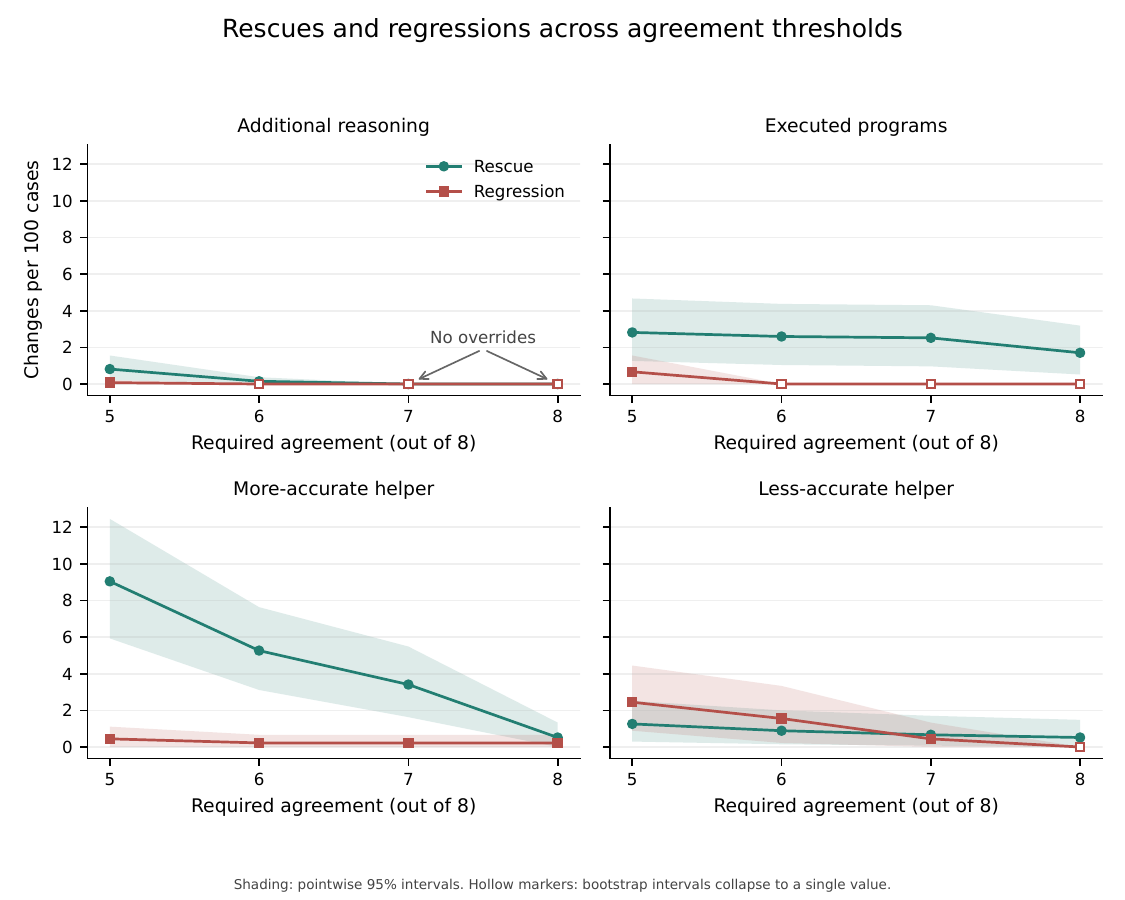}
\caption{\textbf{Rescues and regressions as the agreement requirement increases.}
Points correspond to agreement thresholds $k\in\{5,6,7,8\}$.
Green circles show rescues (wrong-to-right changes) and red squares show regressions (right-to-wrong changes), each per 100 cases.
Panels compare additional same-model reasoning, executed programs, and helpers with higher or lower standalone accuracy than their bases.
Shading shows pointwise 95\% question-bootstrap intervals for rescues and regressions separately.
Labels identify settings with no overrides.
Hollow markers indicate estimates whose bootstrap confidence intervals collapse to a single value.}
\label{fig:fixed16-changes}
\end{figure}

Figure~\ref{fig:fixed16-changes} compares rescues and regressions when an override requires at least $k\in\{5,6,7,8\}$ of the eight additional responses to agree on a different answer.
Additional same-model reasoning produces few rescues or regressions at $k=5$ and $k=6$, and no overrides at $k=7$ or $k=8$.

Executed programs produce more rescues than regressions at every evaluated setting.
Increasing the requirement from $k=5$ to $k=6$ reduces regressions from 0.67 to zero per 100 cases, while rescues decrease only from 2.81 to 2.59.
Requiring $k=7$ or $k=8$ reduces rescues further, while observed regressions remain at zero.
The accuracy gains at $k=5$ and $k=6$ are similar (2.15 and 2.59 percentage points), but only $k=5$ introduces regressions.

Cross-model results differ according to whether the helper is more or less accurate than the base when used alone.
More-accurate helpers produce more rescues than regressions at every setting, but rescues decrease from 9.04 to 0.52 per 100 cases as $k$ increases from 5 to 8.
Less-accurate helpers produce more regressions than rescues at $k=5$ and $k=6$, but more rescues than regressions at $k=7$ and $k=8$.
The observed accuracy changes range from $-1.19$ to $+0.52$ percentage points.
The 95\% confidence interval includes zero at every setting, leaving the direction of the accuracy change uncertain.

Unanimity leaves few overrides even for the methods that still change the vote: programs override on eight distinct questions, more-accurate helpers on five, and less-accurate helpers on two.
We next examine the fractions of overrides that produce rescues or regressions, alongside how often each method overrides the vote. 

\subsection{Stricter agreement does not consistently improve rescue fractions}
\begin{figure}[t]
\centering
\includegraphics[width=\linewidth]{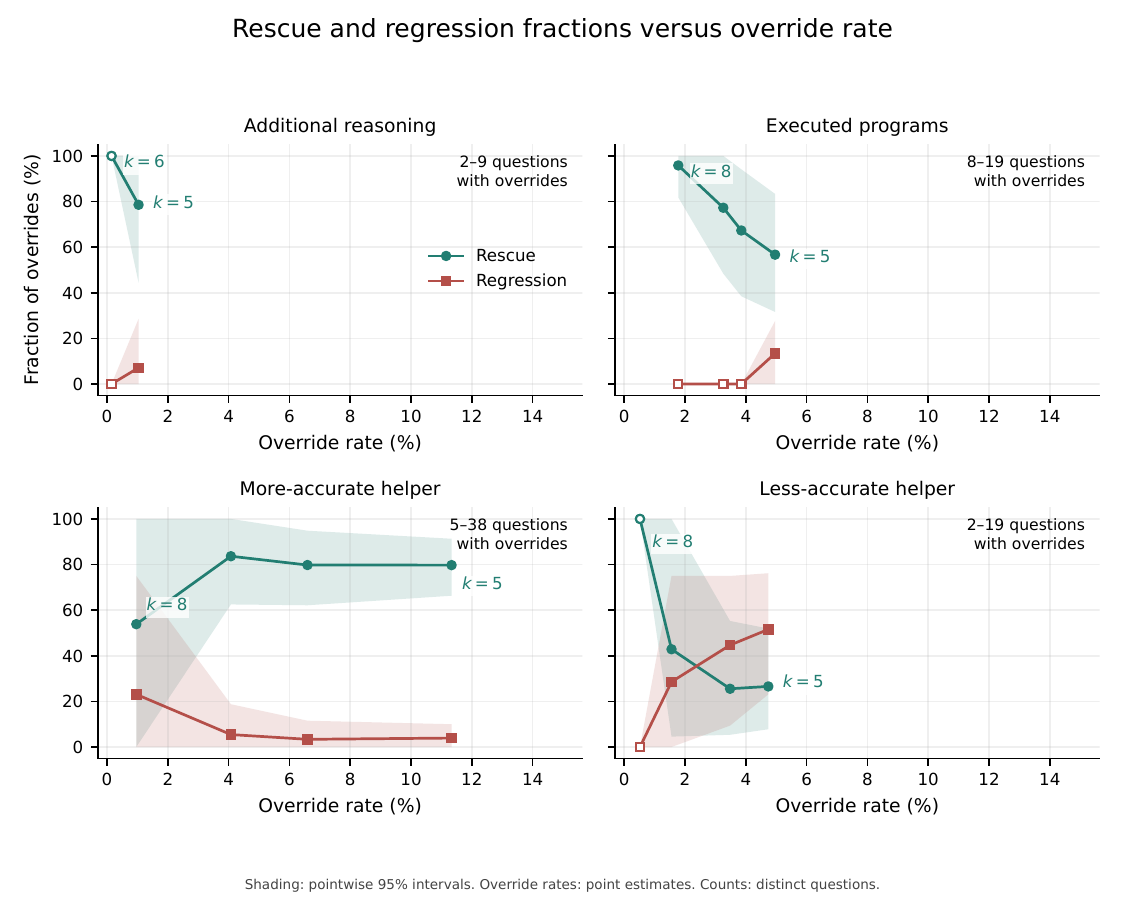}
\caption{\textbf{Rescue and regression fractions versus override rate.}
The horizontal axis shows how often the vote changes; the vertical axis shows the fractions of those overrides that rescue an incorrect answer or replace a correct answer with an incorrect one.
Points correspond to agreement thresholds $k\in\{5,6,7,8\}$.
Endpoint labels identify the displayed range, with larger $k$ toward the left.
Rescue and regression points at the same horizontal position share the same $k$.
Fractions are undefined without overrides, so additional reasoning ends at $k=6$.
Green circles and red squares show rescue and regression fractions, respectively.
Shading shows pointwise 95\% question-bootstrap intervals for the fractions; override rates are point estimates.
Hollow markers indicate collapsed bootstrap intervals.}
\label{fig:fixed16-tradeoff}
\end{figure}

Stricter agreement reduces override frequency but does not consistently increase the fraction of overrides that rescue an incorrect answer (Figure~\ref{fig:fixed16-tradeoff}).
For more-accurate helpers, this fraction decreases from 84\% at $k=7$ to 54\% at $k=8$.
These fractions describe outcomes among overrides, rather than the helper's overall accuracy.
The unanimous overrides occur on only five distinct questions, so the observed decrease does not establish that unanimity generally reduces reliability.

Overrides are rare for different reasons across methods.
At $k=5$, additional same-model reasoning usually confirms the original vote.
When that vote is wrong, the responses most often fail to agree sufficiently on any answer.
Programs are more often limited by too few usable outputs, especially under unanimity.
Appendix~\ref{app:override-reasons} gives the detailed breakdown for each method and agreement requirement.

\subsection{Does combining models outperform the stronger model alone?}
\begin{figure}[t]
\centering
\includegraphics[width=\linewidth]{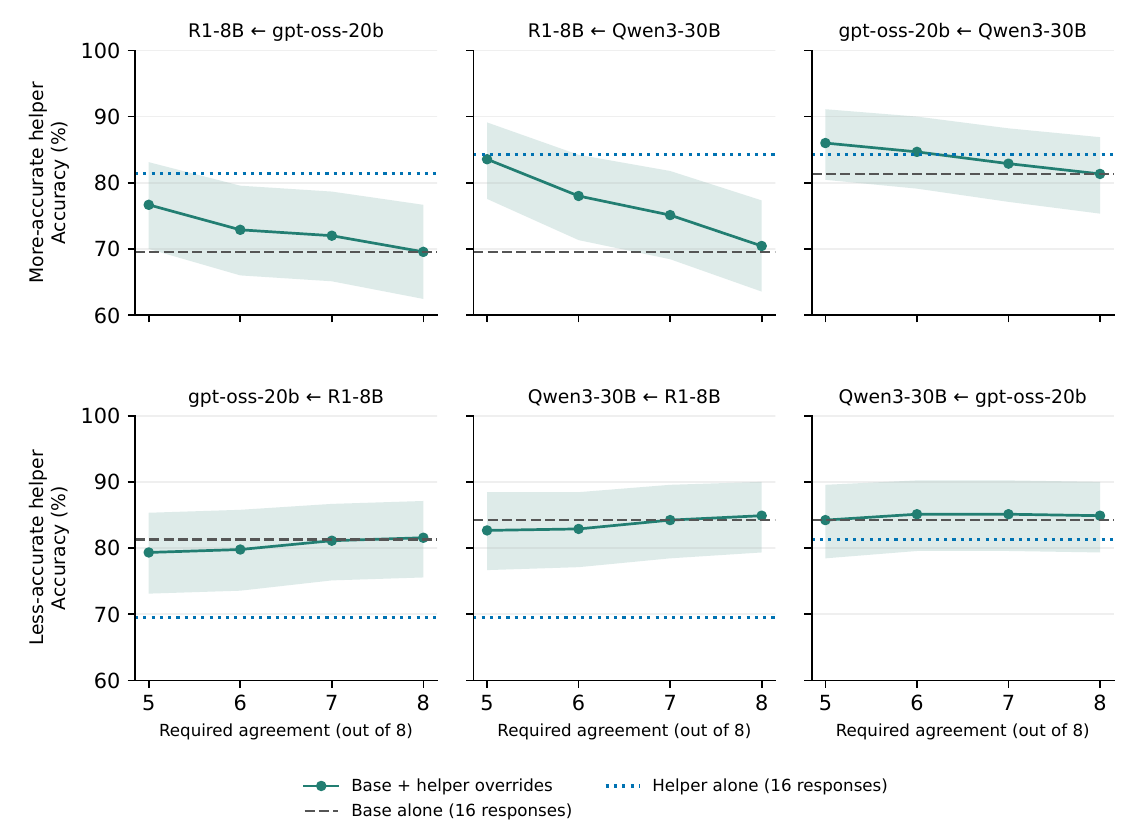}
\caption{\textbf{Combined accuracy compared with both models used alone.}
Each panel shows one base--helper model pair, labeled base $\leftarrow$ helper.
The top row uses more-accurate helpers; the bottom row reverses the model roles.
Green curves show accuracy with cross-model overrides.
Points correspond to agreement thresholds $k\in\{5,6,7,8\}$.
The dashed gray and dotted blue lines show standalone majority-vote accuracy from 16 base and 16 helper responses, respectively.
Shading shows pointwise 95\% question-bootstrap intervals for combined accuracy.
All intervals are unadjusted for multiple comparisons.}
\label{fig:cross-directed}
\end{figure}

Improvement over a weaker base does not by itself justify combining models: the relevant alternative is to use the stronger model alone.
We therefore compare each cross-model method with the more accurate standalone model in its pair (Figure~\ref{fig:cross-directed}).
We show all six model pairs, separating helpers with higher and lower standalone accuracy than the base.
Standalone 16-response voting accuracy increases from R1-8B to gpt-oss-20b to Qwen3-30B.
Here, stronger refers to this measured accuracy, rather than parameter count alone.

Using gpt-oss-20b or Qwen3-30B as a helper increases R1-8B's accuracy at some thresholds, but the combined accuracy remains below that of the helper used alone with 16 responses.
In the reverse comparisons, using R1-8B as a helper decreases the accuracy of both gpt-oss-20b and Qwen3-30B at $k=5$ and $k=6$.
Combining gpt-oss-20b and Qwen3-30B in either direction does not establish an improvement over Qwen3-30B alone: the paired confidence intervals for the accuracy differences include zero at every threshold.

Voting on the stronger helper's eight responses provides a lower-compute alternative to the combined method (Table~\ref{tab:cross-paired}).
It uses exactly the same helper responses while saving the computation needed to generate 16 base-model responses.
Despite this additional computation, combining R1-8B with either stronger helper yields lower observed accuracy than voting on the helper's eight responses alone at every threshold.
Combining gpt-oss-20b with Qwen3-30B does not establish an improvement over Qwen3-30B's eight-response vote.

\section{Related Work}
\label{sec:related}

\paragraph{What evidence justifies changing a majority vote?}
A correct answer can appear among the sampled responses and still lose the majority vote~\citep{wang2022selfconsistency,brown2024monkeys}.
A trained verifier can select among sampled solutions by ranking them~\citep{cobbe2021training}.
ARBITER revisits the majority vote using additional evidence from the same model, including hidden states in one variant~\citep{cai2026arbiter}.
In multi-agent debate, Minority Sentinel learns from the debate record when to accept a minority answer and tracks both correct and incorrect changes~\citep{he2026sentinel}.
In a related analysis, \citet{shu2026pivotal} replace one vote on whether a response is correct with a verifier's verdict and examine cases where that change can decide the outcome.
We start from a majority vote over final answers and compare three ways to obtain an alternative: additional reasoning by the same model, executed programs, and reasoning by a helper model.
Applying the same agreement rule to all three lets us compare how often each method overrides and how many rescues and regressions follow.

\paragraph{Reasoning and program execution.}
A model can retain an error when asked to reconsider its reasoning~\citep{huang2024cannot,stechly2024selfverification}.
Prior work executes model-written programs to answer mathematical questions and uses their outputs to assess sampled solutions~\citep{gao2023pal,chen2023pot,toh2024prove}.
\citet{zhao2023selection} use a language model to choose between a chain-of-thought solution and a program-aided solution.
We compare agreement among program outputs with agreement among additional reasoning responses on the same questions.

\paragraph{When is it worth starting with the weaker model's vote?}
Learning to defer studies when to pass a decision to another decision-maker~\citep{madras2018defer,mozannar2020consistent}.
\citet{aggarwal2023adaptive} use agreement among one model's sampled answers to decide when to stop sampling.
\citet{yue2024large} use agreement among a weaker model's answers to decide which questions warrant the cost of calling a stronger model.
When both models have already supplied answers, a different question arises: does keeping the weaker model's vote as the default improve on voting with the stronger model alone?
We compare the combined answer with each model's standalone majority vote, applying the same comparison used to assess complementarity in human--algorithm collaboration~\citep{donahue2022complementarity}.
When the helper is more accurate, voting on the same eight helper responses used for overrides shows what the weaker model's 16 responses add.
This comparison measures the accuracy gained or lost by using the base vote once the helper responses are available.

\section{Limitations and Conclusion}
\label{sec:limitations}
\label{sec:conclusion}

\paragraph{Program execution.}
Program execution can support useful corrections, but obtaining a correct answer requires formulating a correct solution, translating it faithfully into code, and producing code that runs successfully.
Successful execution establishes that the program ran, rather than that it solves the original problem.
Our comparisons do not separately measure these requirements or isolate execution as the cause of improvement.

\paragraph{Agreement and override frequency.}
With the responses held fixed, increasing the agreement requirement can only remove overrides, so rescues and regressions cannot increase.
Fewer regressions can therefore reflect fewer answer changes rather than more reliable overrides.

\paragraph{Limitations.}
The main comparison covers five mathematical contests, three open reasoning models, and override methods supported by the archived responses.
Selecting questions with the required evidence permits matched comparisons but limits how representative the results are of mathematical reasoning generally.
Repeated baseline sets share correction evidence, and strict agreement leaves overrides on few distinct questions.
The question-bootstrap intervals reflect uncertainty across questions conditional on the observed models, contests, and traces; they do not quantify generalization to new models or tasks.
Appendix~\ref{app:fixed16} documents the choice of question set and the limitations of the archived records.

\paragraph{Conclusion.}
For the evaluated models and questions, our results favor using a more accurate model directly before adding an override method to a weaker model.
Cross-model overrides improve some weaker baselines, but do not establish an advantage over the more accurate model alone, even when its standalone vote uses only eight responses.
Additional same-model reasoning rarely overrides, while executed programs provide useful corrections in some cases.
These findings support comparing override methods with stronger standalone models and reporting rescues, regressions, and override frequency alongside overall accuracy.

\section*{AI use statement}
%%%%%%%%%%%%%%%%%%%%%%%%%%%%%%%%%%%%%%%%%%%%%%%%%%%%%%%%%%%%%%%%%%%%%%%%%%%%%%%

In this work, we used generative AI tools under the authors' direction to assist with implementing the sampler, execution sandbox, and evaluation and statistics scripts; designing and providing feedback on experiments and analyses; proposing and refining hypotheses; interpreting results; and cleaning and reformatting benchmark files.
We have not used generative AI tools to generate synthetic data sets, to formulate or prove mathematical claims beyond the elementary accounting identities in Section~\ref{sec:setting}, or for translation, and qualitative data analysis is not applicable to this work.
Additionally, we used generative AI tools to search for and summarize related literature, format references, suggest manuscript structure, draft and edit text, and implement figures.
Checks of AI-assisted work included comparing reported main-text results with saved analysis outputs, checking descriptions of closely related methods against the cited papers, and inspecting generated figures.
The language models whose outputs are studied (DeepSeek-R1-0528-Qwen3-8B, gpt-oss-20b, Qwen3-30B-A3B-Thinking) are objects of study, not writing tools.
We take responsibility for the final content of this work, including text, claims and artifacts produced with the aid of generative AI.

%%%%%%%%%%%%%%%%%%%%%%%%%%%%%%%%%%%%%%%%%%%%%%%%%%%%%%%%%%%%%%%%%%%%%%%%%%%%%%%
\section*{Ethics statement}

This work evaluates override methods on public competition-mathematics benchmarks, with a separate GPQA Diamond evaluation in Appendix~\ref{app:gpqa}.
It involves no human subjects, personal data or new data sets beyond reformatted public benchmarks, and we see no direct risk of harm.
We report both rescues and regressions to assess how often overrides correct errors or replace correct answers with incorrect ones.

\section*{Reproducibility statement}
%%%%%%%%%%%%%%%%%%%%%%%%%%%%%%%%%%%%%%%%%%%%%%%%%%%%%%%%%%%%%%%%%%%%%%%%%%%%%%%

The fixed-16 analysis is specified in Appendix~\ref{app:fixed16}.
Its compact data retain extracted answers, execution outcomes, source-record provenance, and hashes for detecting reused traces.
The analysis scripts reproduce the main figures, paired contrasts, and broader robustness results.
Prompts, sampling protocols, and experimental history are documented in the appendices.

%%%%%%%%%%%%%%%%%%%%%%%%%%%%%%%%%%%%%%%%%%%%%%%%%%%%%%%%%%%%%%%%%%%%%%%%%%%%%%%
\bibliography{refs}
\bibliographystyle{iclr2027_conference}
%%%%%%%%%%%%%%%%%%%%%%%%%%%%%%%%%%%%%%%%%%%%%%%%%%%%%%%%%%%%%%%%%%%%%%%%%%%%%%%

\appendix
\clearpage
\section{Questions, responses, and statistical analysis}
\label{app:fixed16}
The current analysis uses archived responses from questions originally designated for development or testing.
We report all four agreement thresholds.
Appendix~\ref{app:historical-results} describes the earlier protocol and its use of test results.

\paragraph{Question selection and sampling.}
We used 30 questions from each of five contests, keeping all eligible questions from the three smaller contests and randomly sampling 30 each from SMT and CMIMC.
We chose equal representation by contest after results were available for all 183 eligible questions.
Appendix~\ref{app:selection} compares the two overlapping question sets.
For each model and question, we randomly divided 64 archived reasoning attempts into four sets of 16.
Three sets provided repeated majority-vote baselines.
The fourth provided a standalone comparison and the eight reasoning responses shared across the three baselines.
We used those same eight responses whenever the model served as a helper.
The three baselines also shared one set of eight programs per model and question.
The question list and response assignments are recorded with the data.

\paragraph{Answer processing.}
We normalized answer formatting before grouping and grading responses, without consulting the reference answer.
Missing answers, recorded truncations, and early stops counted as abstentions.
For older R1-8B records without completion metadata, we retained the extracted answer.

\paragraph{Weighting and uncertainty.}
We first averaged the three baseline outcomes for each model and question, or for each base--helper pair and question in cross-model comparisons.
Each model or model pair then received equal weight within its group.
The two helper groups differ in their base models as well as in helper accuracy.
We used 10,000 bootstrap samples of questions within contests, holding the response sets fixed.
For rescue and regression fractions, we excluded resamples with no overrides because the fractions are undefined.

\paragraph{Generation and execution.}
We sampled responses with vLLM at temperature $0.6$ and top-$p=0.95$, with output limits of 60,000 tokens for gpt-oss-20b and 92,000 for the other models.
The gpt-oss-20b chat template used its default reasoning level and included the request for high reasoning in the developer instructions.
Each program ran in Python with a 60-second timeout and a 6 GB memory limit.
We took the answer from the last \texttt{ANSWER:} line of a successful run and excluded programs that only printed a literal.

\paragraph{Reasoning prompt.}
\begin{scriptsize}\begin{verbatim}
Solve the following competition problem. Reason carefully and put your final answer in
\boxed{}.

Problem:
<problem>
\end{verbatim}\end{scriptsize}

\paragraph{Program prompt.}
\begin{scriptsize}\begin{verbatim}
Solve the following competition problem by writing a complete Python program that COMPUTES
the answer.

Rules (a response that violates them is invalid and will be discarded):
- Your final answer MUST be exactly one ```python code block containing a complete, runnable
  program. Stating the answer without such a program, or putting only `ANSWER: <value>` inside
  the block, is invalid.
- The program must compute the value by exhaustive search or enumeration, exact arithmetic
  with fractions or sympy, solving equations, or numerical evaluation with verification. A
  program whose only work is to print a number you derived in your head is invalid. If a
  direct search over the problem's actual parameters is infeasible, compute the answer from a
  formula that the program itself verifies on small cases.
- You cannot run code here (no tool or interpreter is available); never call a tool. Think
  first if you need to.
- Use only the Python standard library, sympy and numpy; it must finish within 60 seconds.
- The last line printed must be exactly `ANSWER: <final answer>` with an exact value (an
  integer, a fraction like 7/3, or a simplified closed form like 3*sqrt(2)+1), not a decimal
  approximation unless the problem asks for one.

Problem:
<problem>
\end{verbatim}\end{scriptsize}

\clearpage
\section{Additional results for the main comparison}
\label{app:additional}
\paragraph{Cross-model controls.}
We evaluate all six model pairs separately.
Standalone helper controls use (i) the exact eight responses supplied to the override method, (ii) the reserved 16-response block containing those eight responses, and (iii) the three disjoint 16-response baseline sets, averaged within question.
The main figure uses the third control and shows the base and helper separately for all six model pairs.
No helper is chosen by its measured performance.
Standalone votes select the most frequent usable answer, with the baseline tie rule and no acceptance threshold.
The standalone 16-response accuracies are 69.6\% for R1-8B, 81.3\% for gpt-oss-20b, and 84.2\% for Qwen3-30B; voting on the eight helper responses gives 71.3\%, 78.0\%, and 84.0\%, respectively.
These controls distinguish relative model accuracy from gains due to combining answers, but do not equalize total inference cost.
Table~\ref{tab:cross-paired} reports paired differences for both directions in Figure~\ref{fig:cross-directed}; all intervals are pointwise and unadjusted for the multiple model pairs and thresholds.

\begin{table}[p]
\centering\small
\setlength{\tabcolsep}{3pt}
\caption{Paired cross-model comparisons on the common 150 questions. Accuracy differences in percentage points with pointwise 95\% paired question-bootstrap intervals. The 16-response controls average baseline runs; the eight-response helper control uses the exact responses supplied to the override method. Intervals are exploratory and unadjusted for multiple comparisons.}\label{tab:cross-paired}
\resizebox{\linewidth}{!}{%
\begin{tabular}{llrccc}
\toprule Base & Helper & Agreement & Versus base (16) & Versus helper (16) & Versus helper (8) \\ \midrule
R1-8B & gpt-oss-20b & 5/8 & +7.11 [+3.56, +11.33] & -4.67 [-9.11, -0.44] & -1.33 [-5.78, +3.11] \\
R1-8B & gpt-oss-20b & 6/8 & +3.33 [+0.89, +6.22] & -8.44 [-13.33, -3.78] & -5.11 [-10.44, +0.00] \\
R1-8B & gpt-oss-20b & 7/8 & +2.44 [+0.44, +5.11] & -9.33 [-14.44, -4.44] & -6.00 [-11.33, -0.67] \\
R1-8B & gpt-oss-20b & 8/8 & +0.00 [+0.00, +0.00] & -11.78 [-17.33, -6.44] & -8.44 [-14.44, -2.89] \\
\addlinespace
R1-8B & Qwen3-30B & 5/8 & +14.00 [+8.89, +19.56] & -0.67 [-4.00, +2.44] & -0.44 [-3.11, +2.22] \\
R1-8B & Qwen3-30B & 6/8 & +8.44 [+4.44, +12.89] & -6.22 [-10.67, -2.00] & -6.00 [-10.44, -1.78] \\
R1-8B & Qwen3-30B & 7/8 & +5.56 [+2.22, +9.33] & -9.11 [-14.22, -4.44] & -8.89 [-14.00, -4.00] \\
R1-8B & Qwen3-30B & 8/8 & +0.89 [+0.00, +2.44] & -13.78 [-19.56, -8.22] & -13.56 [-19.33, -7.78] \\
\addlinespace
gpt-oss-20b & R1-8B & 5/8 & -2.00 [-5.33, +0.89] & +9.78 [+5.33, +14.67] & +8.00 [+3.33, +12.89] \\
gpt-oss-20b & R1-8B & 6/8 & -1.56 [-4.44, +1.11] & +10.22 [+5.56, +15.11] & +8.44 [+3.56, +13.56] \\
gpt-oss-20b & R1-8B & 7/8 & -0.22 [-1.78, +0.89] & +11.56 [+6.44, +16.89] & +9.78 [+4.44, +15.33] \\
gpt-oss-20b & R1-8B & 8/8 & +0.22 [+0.00, +0.67] & +12.00 [+6.67, +17.56] & +10.22 [+4.67, +16.00] \\
\addlinespace
gpt-oss-20b & Qwen3-30B & 5/8 & +4.67 [+1.11, +8.67] & +1.78 [-0.67, +4.44] & +2.00 [-0.44, +4.67] \\
gpt-oss-20b & Qwen3-30B & 6/8 & +3.33 [+0.44, +6.44] & +0.44 [-2.44, +3.33] & +0.67 [-2.89, +4.22] \\
gpt-oss-20b & Qwen3-30B & 7/8 & +1.56 [-0.89, +4.22] & -1.33 [-4.67, +2.00] & -1.11 [-5.11, +2.67] \\
gpt-oss-20b & Qwen3-30B & 8/8 & +0.00 [-1.78, +1.56] & -2.89 [-6.67, +0.89] & -2.67 [-6.89, +1.56] \\
\addlinespace
Qwen3-30B & R1-8B & 5/8 & -1.56 [-4.67, +1.33] & +13.11 [+8.22, +18.22] & +11.33 [+6.22, +16.67] \\
Qwen3-30B & R1-8B & 6/8 & -1.33 [-4.00, +1.33] & +13.33 [+8.22, +18.67] & +11.56 [+6.22, +17.11] \\
Qwen3-30B & R1-8B & 7/8 & +0.00 [-2.00, +2.00] & +14.67 [+9.33, +20.22] & +12.89 [+7.33, +18.67] \\
Qwen3-30B & R1-8B & 8/8 & +0.67 [+0.00, +2.00] & +15.33 [+9.78, +21.11] & +13.56 [+8.00, +19.56] \\
\addlinespace
Qwen3-30B & gpt-oss-20b & 5/8 & -0.00 [-2.44, +2.44] & +2.89 [-0.44, +6.44] & +6.22 [+2.22, +10.44] \\
Qwen3-30B & gpt-oss-20b & 6/8 & +0.89 [+0.00, +2.44] & +3.78 [+0.00, +7.78] & +7.11 [+2.67, +12.00] \\
Qwen3-30B & gpt-oss-20b & 7/8 & +0.89 [+0.00, +2.44] & +3.78 [+0.00, +7.78] & +7.11 [+2.67, +12.00] \\
Qwen3-30B & gpt-oss-20b & 8/8 & +0.67 [+0.00, +2.00] & +3.56 [-0.22, +7.56] & +6.89 [+2.44, +11.78] \\
\addlinespace
\bottomrule
\end{tabular}}
\end{table}

\begin{figure}[t]
\centering
\includegraphics[width=0.82\linewidth]{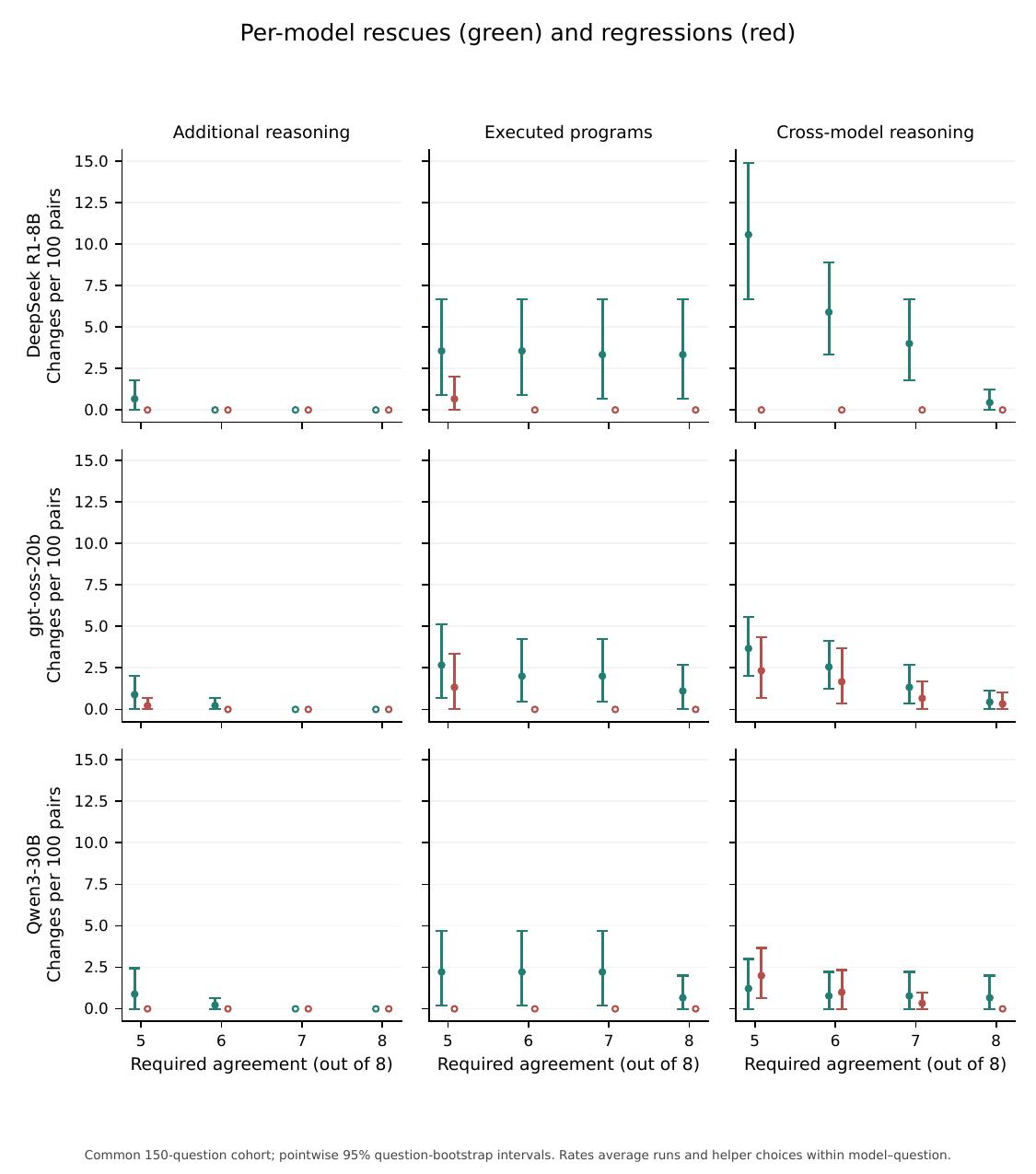}
\caption{Per-model rescues and regressions on the same 150 questions.
Rates average baseline sets and helper choices within model--question pairs.
Intervals are pointwise 95\% question-bootstrap intervals.
These panels expose model variation that the pooled figures average over.}
\label{fig:fixed16-permodel}
\end{figure}

\subsection{Why methods retain the original vote}
\label{app:override-reasons}
Each decision has one of four outcomes.
If fewer than $k$ attempts yield usable answers, we record too few usable answers.
Otherwise, responses either split across answers with fewer than $k$ votes each, agree on the original vote, or agree on an alternative.
Table~\ref{tab:override-reasons-all} reports percentages over all cases.
Table~\ref{tab:override-reasons-incorrect} restricts the denominator to incorrect baseline votes, retaining the same weighting across model pairs and questions.

\begin{table}[p]
\centering\footnotesize
\setlength{\tabcolsep}{3pt}
\caption{Reasons for retaining or overriding the vote on the balanced 150-question set. Entries are percentages with pointwise 95\% question-bootstrap intervals. Same-model outcomes average baseline sets within model--question pairs. Each helper group equally weights its three model pairs after averaging baseline sets within each pair and question. Categories are mutually exclusive and exhaustive; small discrepancies in row sums reflect rounding.}\label{tab:override-reasons-all}
\begin{tabular}{llcccc}
\toprule Procedure & $k$ & Too few usable & Split answers & Confirms vote & Overrides \\ \midrule
Additional reasoning & 5 & 1.6 [0.7, 2.7] & 20.9 [16.4, 25.6] & 76.5 [71.5, 81.3] & 1.0 [0.4, 1.8] \\
 & 6 & 3.1 [1.8, 4.7] & 27.8 [22.7, 33.1] & 69.0 [63.3, 74.6] & 0.1 [0.0, 0.4] \\
 & 7 & 6.0 [4.0, 8.0] & 34.4 [28.9, 40.2] & 59.6 [53.3, 65.8] & 0.0 [0.0, 0.0] \\
 & 8 & 8.7 [6.4, 11.1] & 43.3 [37.6, 49.3] & 48.0 [41.3, 54.7] & 0.0 [0.0, 0.0] \\
\addlinespace
Executed programs & 5 & 18.4 [14.2, 22.9] & 20.4 [16.0, 25.1] & 56.1 [49.8, 62.4] & 5.0 [2.7, 7.5] \\
 & 6 & 25.6 [20.9, 30.2] & 22.2 [18.0, 26.7] & 48.4 [42.1, 54.5] & 3.9 [1.9, 6.0] \\
 & 7 & 36.4 [30.9, 42.0] & 20.9 [16.7, 25.3] & 39.4 [33.3, 45.6] & 3.3 [1.6, 5.3] \\
 & 8 & 54.7 [48.9, 60.2] & 18.0 [14.2, 22.0] & 25.6 [20.4, 30.8] & 1.8 [0.6, 3.3] \\
\addlinespace
More-accurate helper & 5 & 1.6 [0.7, 2.7] & 15.6 [11.6, 20.0] & 71.6 [65.6, 77.3] & 11.3 [7.9, 14.9] \\
 & 6 & 3.1 [1.8, 4.7] & 23.8 [18.7, 29.1] & 66.5 [60.1, 72.7] & 6.6 [4.2, 9.3] \\
 & 7 & 6.0 [4.0, 8.0] & 29.6 [23.8, 35.3] & 60.4 [53.7, 67.0] & 4.1 [2.1, 6.3] \\
 & 8 & 9.1 [6.7, 11.8] & 38.7 [32.7, 44.9] & 51.3 [44.4, 58.0] & 1.0 [0.1, 2.0] \\
\addlinespace
Less-accurate helper & 5 & 1.6 [0.7, 2.7] & 26.2 [20.4, 32.2] & 67.5 [61.0, 73.7] & 4.7 [2.5, 7.2] \\
 & 6 & 3.1 [1.8, 4.7] & 31.8 [25.6, 38.2] & 61.6 [54.9, 68.2] & 3.5 [1.6, 5.6] \\
 & 7 & 6.0 [4.0, 8.0] & 39.3 [32.9, 45.8] & 53.1 [46.1, 60.1] & 1.6 [0.4, 3.1] \\
 & 8 & 8.2 [6.0, 10.4] & 48.0 [41.8, 54.4] & 43.3 [36.3, 50.2] & 0.5 [0.0, 1.5] \\
\addlinespace
\bottomrule
\end{tabular}
\end{table}

\begin{table}[p]
\centering\footnotesize
\setlength{\tabcolsep}{3pt}
\caption{Reasons for retaining or overriding an incorrect vote on the balanced 150-question set. Entries are percentages with pointwise 95\% question-bootstrap intervals. Same-model outcomes average baseline sets within model--question pairs. Each helper group equally weights its three model pairs after averaging baseline sets within each pair and question. Categories are mutually exclusive and exhaustive; small discrepancies in row sums reflect rounding. Rates condition on incorrect baseline votes after within-pair weighting. The all-model baseline errors occur on 57 distinct questions; helper groups can involve different subsets.}\label{tab:override-reasons-incorrect}
\begin{tabular}{llcccc}
\toprule Procedure & $k$ & Too few usable & Split answers & Confirms vote & Overrides \\ \midrule
Additional reasoning & 5 & 4.8 [1.1, 9.0] & 59.9 [50.2, 69.7] & 30.8 [19.8, 41.7] & 4.5 [1.6, 8.0] \\
 & 6 & 8.9 [4.3, 14.0] & 65.8 [55.6, 76.4] & 24.7 [14.4, 35.1] & 0.7 [0.0, 1.9] \\
 & 7 & 14.7 [9.3, 20.1] & 70.9 [61.0, 80.6] & 14.4 [6.2, 23.6] & 0.0 [0.0, 0.0] \\
 & 8 & 15.8 [9.7, 21.9] & 76.0 [67.7, 84.3] & 8.2 [1.9, 16.1] & 0.0 [0.0, 0.0] \\
\addlinespace
Executed programs & 5 & 29.1 [19.1, 39.5] & 40.4 [31.4, 49.5] & 10.6 [3.7, 19.1] & 19.9 [11.6, 29.0] \\
 & 6 & 40.1 [28.3, 52.1] & 35.6 [25.8, 45.4] & 6.5 [0.7, 14.0] & 17.8 [9.5, 27.1] \\
 & 7 & 51.0 [38.7, 63.5] & 31.2 [21.8, 41.1] & 2.7 [0.0, 7.0] & 15.1 [7.4, 23.8] \\
 & 8 & 66.8 [54.5, 78.6] & 22.3 [12.1, 33.7] & 2.7 [0.0, 7.0] & 8.2 [2.9, 14.8] \\
\addlinespace
More-accurate helper & 5 & 4.7 [1.6, 8.5] & 37.7 [27.9, 47.5] & 16.5 [8.3, 25.4] & 41.1 [31.0, 51.7] \\
 & 6 & 8.9 [4.6, 13.5] & 54.5 [45.2, 63.5] & 12.6 [5.2, 20.8] & 24.0 [16.1, 32.8] \\
 & 7 & 15.6 [10.3, 20.9] & 59.8 [50.0, 69.4] & 10.1 [3.3, 17.6] & 14.5 [7.8, 22.2] \\
 & 8 & 20.7 [14.0, 27.7] & 69.0 [59.9, 77.5] & 7.5 [1.7, 14.7] & 2.8 [0.3, 6.3] \\
\addlinespace
Less-accurate helper & 5 & 3.1 [0.0, 7.1] & 61.5 [47.1, 76.2] & 21.7 [8.7, 35.1] & 13.7 [5.9, 23.2] \\
 & 6 & 7.1 [2.3, 12.4] & 61.1 [46.3, 76.1] & 20.4 [8.3, 32.9] & 11.5 [4.3, 20.4] \\
 & 7 & 16.8 [10.3, 23.0] & 64.6 [52.2, 76.7] & 11.9 [2.8, 22.8] & 6.6 [1.4, 13.7] \\
 & 8 & 16.8 [10.3, 23.0] & 73.5 [63.6, 82.9] & 6.6 [0.0, 16.6] & 3.1 [0.0, 9.5] \\
\addlinespace
\bottomrule
\end{tabular}
\end{table}

\section{Sensitivity to question selection}
\label{app:selection}
The full eligible set adds 23 SMT and 10 CMIMC questions to the balanced set, keeping the trace partitions and evidence for the original 150 questions.
Table~\ref{tab:balanced-robustness} compares accuracy changes at each threshold.
Under unanimity, the program gain is 1.70 percentage points on the balanced set and 1.64 on the full set.
The cross-model rows average over both helpers for each base and therefore assess only that aggregate.
The two question sets overlap.

\begin{table}[p]
\centering\small
\caption{Sensitivity to balanced contest sampling with identical trace partitions for shared questions. Net accuracy changes in percentage points with pointwise 95\% paired question-bootstrap intervals. All thresholds are reported. A dash denotes a degenerate interval, not established zero risk.}\label{tab:balanced-robustness}
\begin{tabular}{lrcc}
\toprule Procedure & Agreement & Balanced 150 & Full 183 \\ \midrule
Additional reasoning & 5/8 & +0.74 [+0.15, +1.48] & +0.67 [+0.12, +1.34] \\
Additional reasoning & 6/8 & +0.15 [+0.00, +0.37] & +0.12 [+0.00, +0.30] \\
Additional reasoning & 7/8 & +0.00 [---] & +0.00 [---] \\
Additional reasoning & 8/8 & +0.00 [---] & +0.00 [---] \\
\addlinespace
Executed programs & 5/8 & +2.15 [+0.37, +4.15] & +2.49 [+0.73, +4.43] \\
Executed programs & 6/8 & +2.59 [+1.04, +4.37] & +2.61 [+1.15, +4.25] \\
Executed programs & 7/8 & +2.52 [+0.96, +4.30] & +2.31 [+0.97, +3.83] \\
Executed programs & 8/8 & +1.70 [+0.52, +3.19] & +1.64 [+0.61, +2.91] \\
\addlinespace
Cross-model reasoning & 5/8 & +3.70 [+2.19, +5.30] & +3.55 [+2.25, +4.95] \\
Cross-model reasoning & 6/8 & +2.19 [+0.93, +3.48] & +2.43 [+1.31, +3.58] \\
Cross-model reasoning & 7/8 & +1.70 [+0.67, +2.85] & +1.82 [+0.88, +2.88] \\
Cross-model reasoning & 8/8 & +0.41 [-0.11, +1.07] & +0.43 [-0.06, +1.00] \\
\addlinespace
\bottomrule
\end{tabular}
\end{table}

\section{Additional evaluations under the original protocol}
\label{app:historical-results}
The original protocol sampled 16 responses and expanded to 64 when agreement was below 80\%.
Correction methods were applied to pools that still had multiple answers and less than 80\% agreement.
For eight-response methods, the reported rule required at least six usable responses and a unique leading answer with share at least $\theta$ among usable responses.
These denominators differ from the eight attempts in the main analysis.

The original primary program rule required unanimity and restricted the answer to the five leading candidates.
The unrestricted unanimous rule was selected on R1-8B pools whose test contests had been used in earlier experiments, then fixed before scoring the other two models.
The strict program prompt also preceded scoring those models.
The lower program thresholds and question-level statistical summaries were explored after viewing test results.
Later evaluations included additional contests after the rules were fixed.

\paragraph{Candidate verification and stronger judges.}
The base model judged each of the five leading answers four times.
When all positive verdicts supported one candidate, verification overrode 12 pools, six on the strength of a single positive verdict.
A stricter check requiring unanimous rejection of the original answer and unanimous acceptance of one alternative gained 0.1 percentage points on each historical split.
On the R1-8B test pools, gpt-oss-120b selected among the five candidates after seeing the problem and candidate list.
At $\theta=0.75$, this increased accuracy by 4.9 percentage points (95\% CI [1.7, 8.9]); the observed change on development was $-1.1$ points.

\paragraph{Executing programs or predicting their outputs.}
For gpt-oss-20b, the matched control used the same generated programs and either executed them or asked the model to predict their outputs.
At $\theta=0.5$ on the historical test set, execution produced 15 overrides on five distinct questions and increased accuracy by 3.3 percentage points (95\% CI [0.8, 6.5]).
Predicted outputs produced no test overrides at that threshold.
The comparison concerns those programs and few questions.

\paragraph{GPQA Diamond.}
\label{app:gpqa}
R1-8B was evaluated on 198 four-option questions with four sampling runs per question.
Of the 792 pools, 353 qualified for correction.
When additional solutions were shown the candidates and unanimously favored an alternative, all 13 observed overrides were rescues.
These are counts of sampling pools; multiple pools can represent the same question.
With only four options, availability among the five leading answers has a different interpretation from availability on open-answer mathematics questions.

\end{document}